\documentclass[letterpaper, 10 pt, conference]{ieeeconf} 

\IEEEoverridecommandlockouts
\usepackage{subcaption}
\usepackage{amsmath,amssymb,amsfonts}
\usepackage{algorithmic}
\usepackage{graphicx}

\begin{document}

\title{\Large \bf
Vision based UAV Navigation through Narrow Passages}

% \author{
%     \IEEEauthorblockN{
%       Jayakant Kumar$^{1}$\thanks{$^{1}$School of Robotics,
%         Defence Institute of Advanced Technology,
%         Pune, India
%         jayakantkumar1470@gmail.com, 
%        pooja\_agrawal@diat.ac.in}
%       Himanshu$^{2}$\thanks{$^{2}$Robotics Research Center, IIIT-H,
%         Hyderabad, India
%         himanshukmr234@gmail.com, 
%         harikumar.k@iiit.ac.in}
%       Harikumar Kandath$^{2}$
%       Pooja Agrawal$^{1}$}
% }

\author{Jayakant Kumar$^{1}$ Himanshu$^{2}$ Harikumar Kandath$^{2}$ Pooja Agrawal$^{1}$% <-this % stops a space
% \thanks{*This work was not supported by any organization}% <-this % stops a space
\thanks{$^{1}$School of Robotics,
        Defence Institute of Advanced Technology,
        Pune, India,
        jayakantkumar1470@gmail.com, 
       pooja\_agrawal@diat.ac.in.}%
\thanks{$^{2}$Robotics Research Center, IIIT-H,
        Hyderabad, India,
        himanshukmr234@gmail.com, 
        harikumar.k@iiit.ac.in.}
\thanks{youtu.be/iN5ixNguQxc}}%

\maketitle

\begin{abstract}
This research paper presents a novel approach for navigating a micro UAV (Unmanned Aerial Vehicle) through narrow passages using only its onboard camera feed and a PID control system. The proposed method uses edge detection and homography techniques to extract the key features of the passage from the camera feed and then employs a tuned PID controller to guide the UAV through and out of the passage while avoiding collisions with the walls. To evaluate the effectiveness of the proposed approach, a series of experiments were conducted using a micro-UAV navigating in and out of a custom-built test environment (constrained rectangular box). The results demonstrate that the system is able to successfully guide the UAV through the passages while avoiding collisions with the walls.

\end{abstract}

% \begin{IEEEkeywords}
% UAV, Computer Vision, 
% \end{IEEEkeywords}

\section{Introduction}
Unmanned aerial vehicles (UAVs) have gained significant attention in recent years due to their wide range of applications in various fields, including surveillance, search and rescue, infrastructure inspection\cite{TAVASOLI2023106193}, and agriculture. However, one of the major challenges in using UAVs is their ability to navigate through confined and narrow spaces. Vision-based navigation systems offer a promising solution to this challenge, using onboard cameras to capture visual information and processing it to guide the UAV through the environment\cite{5980136}.

This paper focuses on the problem of vision-based UAV navigation through narrow passages. This problem is particularly challenging due to the tight space constraints and the need for accurate sensing, localization, and obstacle avoidance. Various methods have been proposed for this problem, including depth sensing\cite{7440994}, lidar\cite{8979150}, stereo vision\cite{9225398}, and visual odometry\cite{9274825}. However, these methods often rely on expensive sensors and are not  suitable for micro UAVs. Also, the existing deep learning methods are computationally expensive to be implemented in a micro UAV \cite{nguyen2020autonomous}.

To address these challenges, we propose a simple and cost-effective approach for navigating a micro UAV through narrow passages using only its onboard camera feed and a PID control system. The proposed approach utilizes edge detection and homography techniques to extract the key features of the passage from the camera feed and employs a tuned PID controller to guide the UAV through the passage while avoiding collisions with the walls.

The main contribution of this paper is the development and evaluation of the proposed homography-based algorithm, which offers several advantages over existing methods for vision-based UAV navigation through narrow passages. Our algorithm can estimate the center point of the opening of the rectangular box when presented with a partial image of the opening area of the box. Without using our method, the UAV collides with either the side wall or the ceiling wall of the box due to sidewall and ceiling effect\cite{sidewallinfluence}. Our method uses the estimate of the opening window's center point to adjust the UAV's position and fly out when the error is below some threshold. The approach is simple, cost-effective, and flexible, making it suitable for various applications and structured environments. 
Overall, the results of this research demonstrate the potential of vision-based navigation systems for overcoming the challenges of navigating through narrow passages using UAVs.

The remainder of the paper is structured: Section II provides the necessary background information and formulates the problem. Section III presents the proposed methodology. Section IV describes our experimental setup and presents the results obtained from evaluating our method. Finally, Section V concludes the paper with a summary of the key findings and an outlook on future research directions.

\begin{figure*}[h!]
    \centering
    \includegraphics[width=0.8\textwidth]{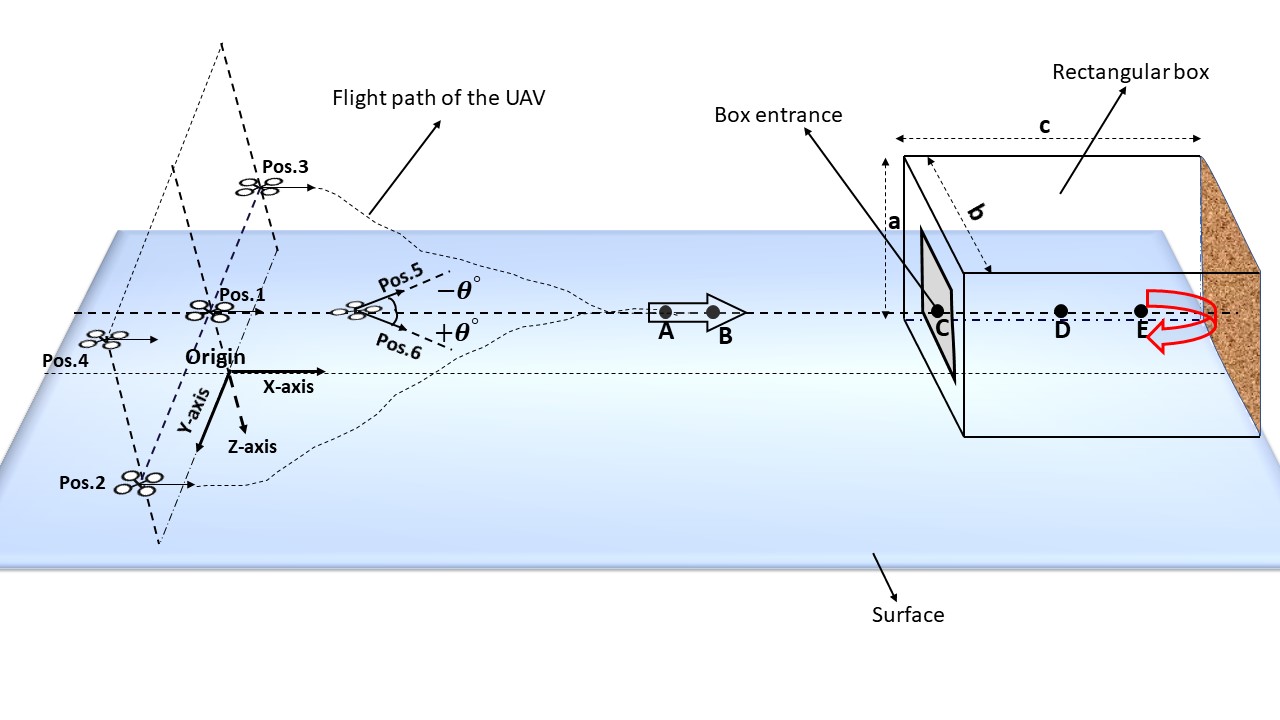}
    \caption{Diagram of the experimental condition. The left side of the diagram shows the different initial positions of the UAV for different flight trials. The right side of the diagram shows the rectangular box used as the constrained space for the UAV to enter and exit.}
    \label{fig:Exp}
\end{figure*}

\begin{figure}[h!]
    \centering
    \includegraphics[width=0.25\textwidth]{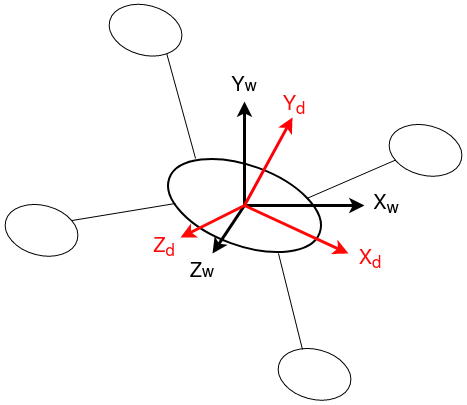}
    \caption{ UAV body frame (RED) and world frame (BLACK) with origin  aligned with UAV's center of gravity.}
    \label{fig:UAV's frame}
\end{figure}

\section{Preliminaries and Problem Formulation}
The problem we aim to address in this study is the navigation of micro UAVs through constraint spaces. Specifically, we focus on flying a quadcopter inside a small rectangular box with a small entrance close to $3$ times the size of the quadcopter as depicted in Fig \ref{fig:Exp}. The challenge lies in safely navigating the UAV through the small opening without collision and stabilizing it inside the box despite its limited dimensions and aerodynamic effects such as sidewall and ceiling effects \cite{sidewallinfluence}. Once the UAV has reached a certain desired distance inside the box, it must exit through the same opening while avoiding collision with the box. The key assumption here is that the width of the entrance and the narrow passage is known prior. This is a reasonable assumption while inspecting structured environments like aircraft cabins, fuel tanks, etc.

The algorithm discussed in this paper uses visual feedback to adjust the UAV's position. We transform the position of an object from the image frame to the UAV frame. At first, we used the pinhole camera model to transform the object's position from the image frame to the camera frame. The equation describing the pinhole camera model is given in Eqn. \ref{eq: xi2w}.

\begin{align}
\label{eq: xi2w}
x_{c} = \frac{D \times x_{i}}{f_{x}}, y_{c} = \frac{D \times y_{i}}{f_{y}}
\end{align}

Where ($x_{c}$, $y_{c}$) and ($x_{i}$,$y_{i}$) are the x and y component of the position of the object in the camera frame and its projection in the image frame. Also, $D$ is the object's distance from the camera, and $f_{x}$ \& $f_{y}$ are the camera's focal length in the x and y directions. \par

After that, ($x_{c}$, $y_{c}$) is rotated from the camera frame to the UAV frame using the Rotation matrix $R_{C}^U$ as mentioned in Eqn. \ref{eq: Transform Rotate_x C2W}. The first step in this rotation is transforming the coordinates from the camera frame to the world frame, which has the origin mounted on the center of gravity of the UAV, as shown in Fig. \ref{fig:UAV's frame}. Next, the coordinates are rotated from the world frame to the UAV frame using the rotation matrices $R(\theta)$ and $R(\phi)$. 

\begin{equation}
\label{eq: Rotate_x C2D}
R_{C}^U= R(\theta)R(\phi)R_{C}^W
\end{equation}

\begin{equation}
\label{eq: Transform Rotate_x C2W}
\begin{bmatrix} X_{u}\\ Y_{u}\\ Z_{u}\end{bmatrix}= R_{C}^U\begin{bmatrix} x_{c} \\ y_{c} \\ 1\end{bmatrix}
\end{equation}

Where $(X_{u}, Y_{u}, Z_{u})$ is the position of the object in the UAV reference frame, $R(\phi)$ and $R(\theta)$ are the rotation matrices along x and y-axis of the UAV frame and $R_{C}^W$, is the rotation matrix for converting the camera frame to world frame. Also, $\phi$ and $\theta$ are the roll and pitch angle of the UAV.\\

The first objective when navigating the UAV in and out of the box is to use the onboard camera’s image to detect the entrance window. The second objective is navigating the UAV through the entrance while avoiding collisions with the walls. The following sections of this paper explain how we accomplish both objectives while going in and out of the box.

\section{Methodology}

This section explains the novel control algorithm used to fly the micro UAV in and out of the rectangular box. The control algorithm consists of two parts. The first part (mentioned in Fig \ref{fig:Flow chart 1}) is responsible for flying the UAV inside the box. The second part (shown in Fig \ref{fig:Flow chart 2}) is used for flying the UAV out of the rectangular box with only the partial feedback from the  image captured using UAV's camera.

\subsection{Part 1: Entering the narrow passage}
\begin{figure}[h!]
    \centering
    \includegraphics[width=0.45\textwidth]{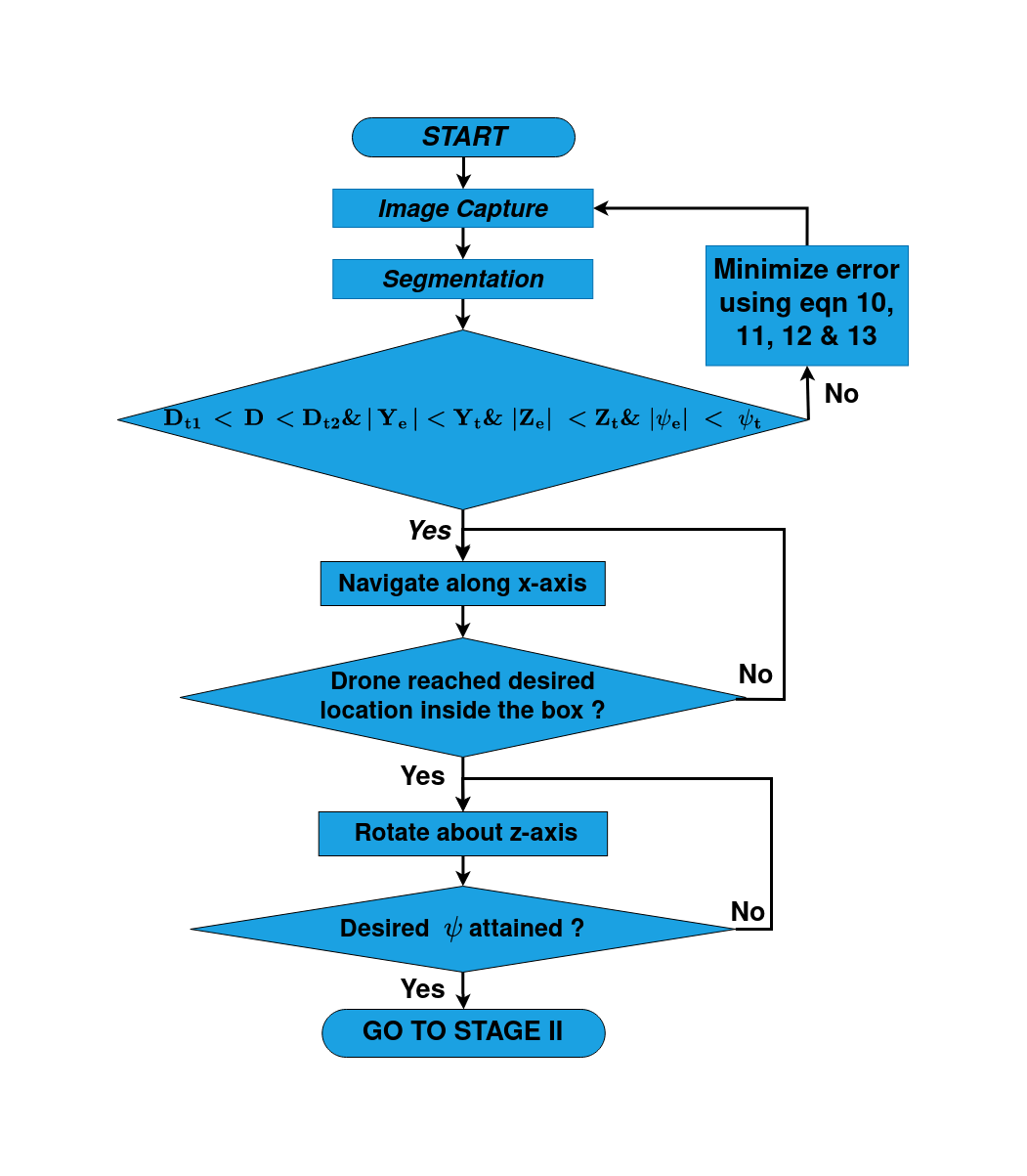}
    \caption{Part 1: Flowchart showing the methodology for UAV entering inside the narrow passage. }
    \label{fig:Flow chart 1}
\end{figure}

In this part, We present our control algorithm for piloting a UAV inside a small rectangular box (shown in Fig \ref{fig:Flow chart 1}). Our approach involves extracting an image from the UAV’s onboard camera and segmenting it to find contours. We then detect the box entrance's corners, center point, and depth and use this information to calculate control commands. This section provides a more detailed explanation of each step in the flow chart.

\subsubsection{Image Capture}
Our method uses image feedback to align the UAV in front of the target. Figure \ref{fig:Front view} shows the schematic diagram of the image captured by the onboard camera. The dimension of the captured image is $M$x$N$ which contains the features of the box entrance. The algorithm later uses these features to accurately determine the image's corner position, center point, and depth.  
\subsubsection{Segmentation}
After capturing the image from the UAV's camera, it is processed to find the contours of the target window. This happens in two steps. First, we pass the grayscale version of the current image through the Canny filter \cite{canny1986computational} for edge detection. Next, we find the boundary coordinates of different contours detected by the canny filter. This is done using the border following algorithm from \cite{findContours}.  \par
The filter detects many contours associated with different objects from the given image. Thus we also remove the irrelevant contours from the set of detected contours. We select the contour whose dimension matches the dimension of the target and reject the others. Next, the center coordinate is detected by finding image moments which are just a weighted average of image pixel intensities. Afterward, the corner points are detected by the Good Features To Track (GFTT) algorithm \cite{goodFeaturestoTrack}.

\subsubsection{Navigation and Control}
To explain the control algorithm, first assume that the reference point is $(x_{ri}, y_{ri})$, the center point of the target is $(x_{pi}, y_{pi})$ in the image frame. Then the positional error in the image frame can be written as shown in Eqn. \ref{eq: x_ep}.

\begin{equation}
\label{eq: x_ep}
x_{ei} = (x_{pi} - x_{ri}),\,\, y_{ei} = (y_{pi} - y_{ri})
\end{equation}

where, $x_{ei}$, $y_{ei}$ are the x and y component of the error in image frame. \\
To change the reference frame of the errors  $x_{ei}$, $y_{ei}$ from the image frame to the UAV frame, we first use the pinhole camera model to get the error in the camera reference frame as given in eqn. \ref{eq: X_ec}.

\begin{equation}
\label{eq: X_ec}
X_{ec} = \frac{D  \times x_{ei}}{f_{x}}, Y_{ec} = \frac{D  \times y_{ei}}{f_{y}}
\end{equation}

where ($X_{ec}$, $Y_{ec}$) are the x and y component of error in camera frame.\\  
Next error in the camera frame is rotated to the UAV frame using the rotation matrix $R_C^U$ defined earlier in the preliminary section as mentioned in the Eqn. \ref{eq: error C2D}.

\begin{equation}
\label{eq: error C2D}
\begin{bmatrix} X_e \\ Y_e \\ Z_e \end{bmatrix}=  
R_C^U\begin{bmatrix} X_{ec} \\ Y_{ec} \\ 1 \end{bmatrix}
\end{equation}
where $X_e$, $ Y_e$ \& $ Z_e$ are the x, y, and z components of the error in the UAV frame.

The pinhole hole camera method also requires the depth or distance of the object from the camera. We calculate the depth $D$ of the object using the Eqn. \ref{eq: DepthCalc} given below.
\begin{equation}
\label{eq: DepthCalc}
D = \frac{f_x\times W_r}{W_i}
\end{equation}
where the $f_{x}$ is the focal length of the camera in the x direction, $W_{r}$ and $W_{i}$ are the actual width and apparent width (in the image frame) of the target window.
\begin{figure}[h!]
    \centering
    \includegraphics[width=0.40\textwidth]{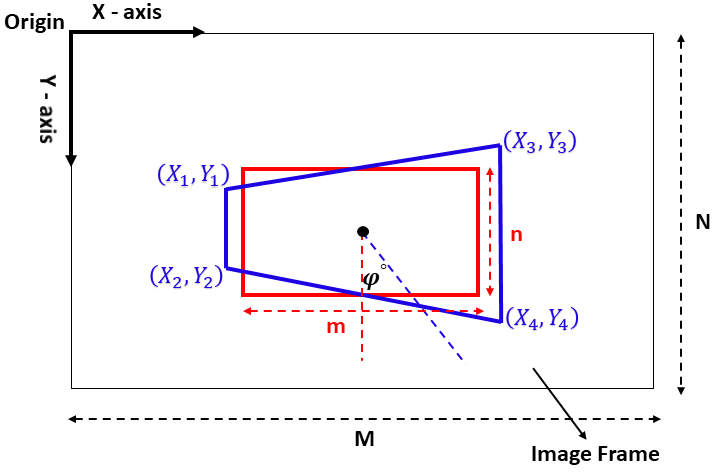}
    \caption{Image frame when UAV is with and without the yaw error.}
    \label{fig:Front view}
\end{figure}

Apart from the positional error, we also calculate the yaw error to adjust the yaw of the UAV during the flight. Fig \ref{fig:Front view} shows the image frame in two cases. If we assume the box's entrance be a plane and the line perpendicular to this plane is denoted by $X_{box}$. When the UAV is in front of the target, i.e., box entrance, the UAV's x-axis aligns with the line $X_{box}$. Thus, the red rectangle is shown in the image frame. However, when the UAV is not aligned with the box's entrance, i.e., the UAV's x-axis is making some angle with $X_{box}$, then the image frame will have the blue trapezoid as shown in Fig \ref{fig:Front view}. In the latter case, we can use the difference between the length of the parallel sides of the trapezoid to create an error proportional to the yaw error, as mentioned in Eqn. \ref{eq: yaw_err}.

\begin{equation}
\label{eq: yaw_err}
\psi_e = (Y_4-Y_3) - (Y_2 -Y_1) 
\end{equation}
where 
$\psi_e$ is the yaw error of the UAV in pixel, and $Y_1, Y_2, Y_3 $ and $Y_4 $ are the y coordinates of the trapezoid's corner points. \par

In the paragraphs before this, we have defined the positional error ($X_{e}, Y_{e}, Z_{e}$), yaw error ($\psi_{e}$), and depth $(D)$. Now, these errors and depth are used to calculate control commands for the UAV. Before applying the control command to the UAV, we check whether these errors are inside a desired threshold as given in Eqn. \ref{eq: Part 1 threshold}. If they  satisfy the condition, the UAV navigates along the positive x-axis. Otherwise, the velocity commands are calculated based on the positional and yaw errors, as shown in Eqn. \ref{eq: X PID}, Eqn. \ref{eq: Y PID}, Eqn. \ref{eq: Z PID} and Eqn. \ref{eq: yaw PID}. 

\begin{align}
\label{eq: Part 1 threshold}
D \in (D_{t1}, D_{t2}), \,\, |Y_{e}|<Y_{t}, \,\, |Z_{e}|<Z_{t}, \,\, |\psi_{e}|<\psi_{t} 
\end{align}

\begin{equation}
\label{eq: X PID}
V_{x} = C
\end{equation}

\begin{equation}
\label{eq: Y PID}
V_{y} = k_{py}  Y_{e} + k_{iy} \int Y_{e}dt + k_{dy} \frac{d Y_{e}}{dt}
\end{equation}

\begin{equation}
\label{eq: Z PID}
V_{z} = k_{pz}  Z_{e} + k_{iz} \int Z_{e}dt + k_{dz} \frac{d Z_{e}}{dt}
\end{equation}

\begin{equation}
\label{eq: yaw PID}
\omega_{z} = k_{p\psi} \psi_{e} + k_{i\psi} \int \psi_{e}dt + k_{d\psi} \frac{d \psi_{e}}{dt}
\end{equation}

where ($V_{x}$, $V_{y}$, $V_{z}$) are the linear velocity command in x, y and z directions. $\omega_{z}$ is the angular velocity command around z axis .($k_{py}, k_{iy}, k_{dy}$), ($k_{pz}, k_{iz}, k_{dz}$) \& ($k_{p\psi}, k_{i\psi}, k_{d\psi}$) are the PID controllers gains. $C$ is the constant velocity command in the x direction.\\
Equation \ref{eq: X PID} shows that a constant velocity command $C$ was used to control the UAV in X direction.  This is because the UAV only needs to reach a certain depth near the target. Once the UAV arrives at this depth, it minimizes $Y_{e}$, $Z_{e}$, and $\psi_{e}$ using the controllers mentioned in Eqn. \ref{eq: Y PID}, \ref{eq: Z PID} \& \ref{eq: yaw PID}. When the errors are below their respective thresholds, the UAV navigates toward a waypoint inside the box using the feedback from its odometry. When the UAV reaches the assigned waypoint inside the box, it stops and rotates until a desired yaw is reached. In our case, we set this desired yaw such that the UAV faces the exit of the box. This is where part 2 of the controller is activated.

\subsection{Part 2: Exiting the narrow passage}
\begin{figure}[h!]
    \centering
    \includegraphics[width=0.45\textwidth]{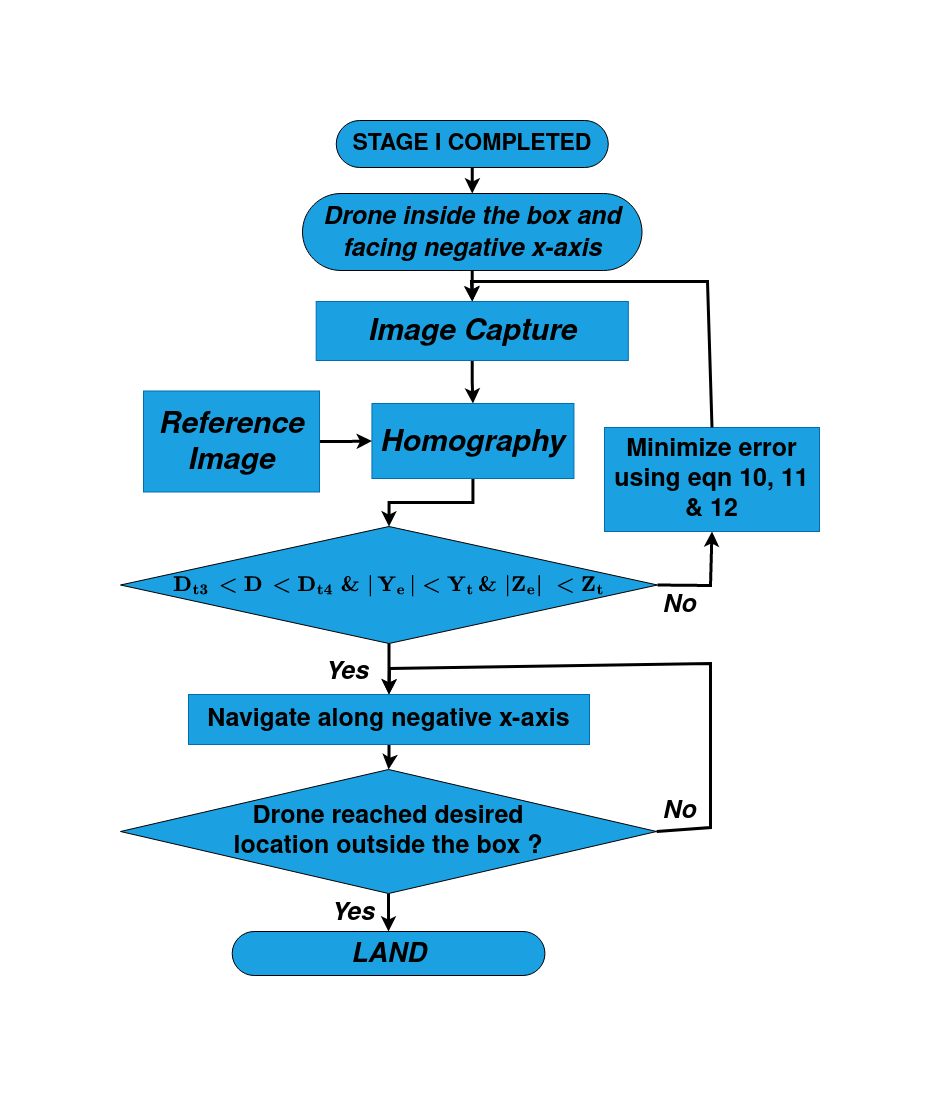}
    \caption{Part 2: Flowchart showing the methodology for UAV coming out of the narrow passage.}
    \label{fig:Flow chart 2}
\end{figure}

The control algorithm for flying the UAV out of a box uses Homography \cite{homography} to estimate the center coordinate of the exit window with partial images. The positional error is calculated by comparing the estimated center coordinate and reference point. Control commands are then calculated using a PID controller to adjust the UAV’s position. The rest of this section provides a more detailed explanation of each step of the algorithm described in Fig. \ref{fig:Flow chart 2}.

\subsubsection{Homography}
Homography \cite{homography} is the technique used in computer vision to describe the relationship between two images of the same object or scene. This technique is specifically useful for cases where the object is occluded or partially visible.  In our case, capturing a full image of the box's exit without collision is difficult due to the box's small size and aerodynamic effects. Homograpy starts by taking a reference image and the image from the UAV's camera. Then features from the reference and current images are extracted using the Sift algorithm \cite{sift}. These extracted features are then matched to form pairs using the kNN algorithm \cite{kNN}. After that, the feature points (from the reference image and current image) with less distance are selected, and the rest are discarded. From the remaining points of the reference and current images, the homography matrix is calculated by the RANSAC algorithm \cite{findHomo}. This homography matrix is then used to transform the feature points of the reference image to be in the current image. Thus, we get the estimates of target contours even when the target is partially visible. After detecting the target contour, the center coordinate is detected by finding image moments. 

\subsubsection{Navigation and Control}
After we get the center coordinates from the homography technique, the steps mentioned in the Navigation and Control section are followed to find the positional errors ($Y_{e}$ \& $Z_{e}$), target depth $D$. We again check for thresholds of position error and depth as mentioned in Equation \ref{eq: Threshold 2}. If the position error does not satisfy Equation \ref{eq: Threshold 2}, velocity commands are generated based on the PID logic given in Equation \ref{eq: X PID}. If all the thresholds are satisfied, the UAV navigates along the negative x-axis to reach the waypoint outside the box using the feedback from its odometry data.

\begin{align}
\label{eq: Threshold 2}
D \in (D_{t3}, D_{t4}), \,\, |Y_{e}|<Y_{t}, \,\, |Z_{e}|<Z_{t} 
\end{align}

\section{Experimental Setup and Results}
This section describes the experimental setup used to test the proposed algorithm and the results obtained. DJI Tello UAV is used for experiments with a front-facing camera. The image size is $480$ x $360$ pixels. Our algorithm runs 5-7 frames per second (fps) on a laptop with $8^{th}$ gen Intel Core i7 processor running at $2.3$ GHz, $16$ GB of RAM, and Ubuntu 20.04 Operating system. \par
\begin{figure}[h!]
    \centering
    \includegraphics[width=0.4\textwidth]{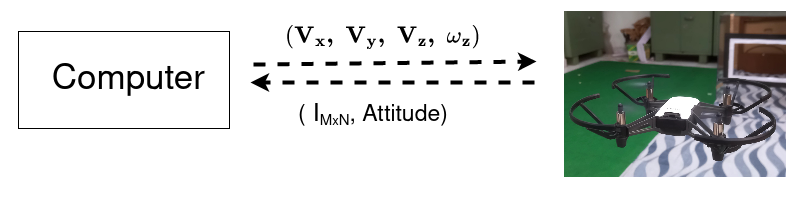}
    \caption{ Communication between Computer and micro UAV}
    \label{fig:Mini UAV}
\end{figure}

\begin{figure}[h!]
    \centering
    \includegraphics[width=0.25\textwidth]{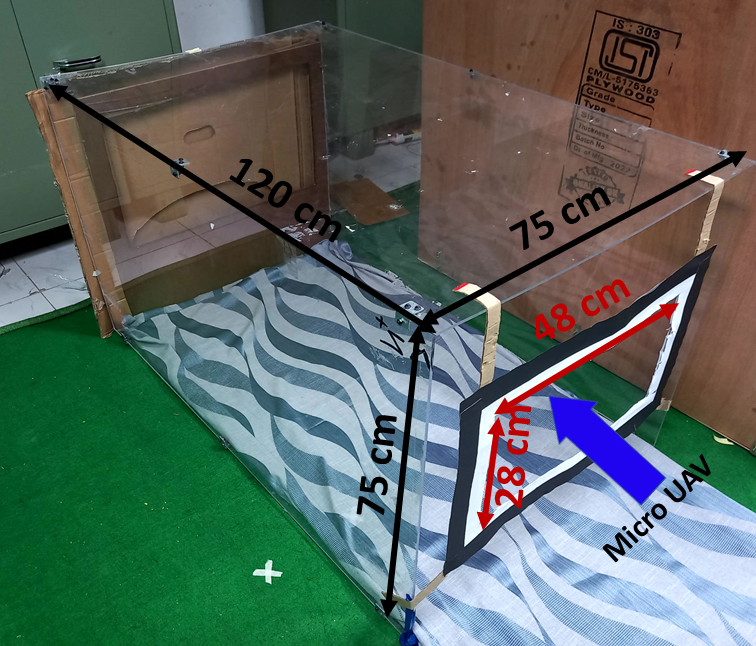}
    \caption{ Rectangular box used in the experiments.}
    \label{fig:Box Dimension}
\end{figure}

The UAV's camera was calibrated using Zhang's method \cite{CameraCalibzhang}. The resulting camera calibration matrix $K$ is given in Equation \ref{eq: caliberationMat}. As seen from the calibration matrix, the focal lengths ($f_{x}$ \& $f_{y}$) of the camera are  $466$ \& $467$ pixels, respectively.

\begin{equation}
\label{eq: caliberationMat}
K = \begin{bmatrix}  466 & 0 & 247 \\ 0 & 467 & 174 \\ 0 & 0 & 1 \end{bmatrix}
\end{equation}

We used a rectangular box of $75$ cm x $75$cm x $120$cm with an entrance window of $48$cm x $28$cm as the narrow passage. Meanwhile, the dimension of the quadcopter used is $17$cm x $17$cm.\par 

While calculating the error, we use some threshold values below, which we consider the UAV stabilized. These thresholds were $D_{t1}= 1.55$m, $D_{t2}= 1.65$m, $D_{t3}= 0.7$m, $D_{t4}= 0.95$m, $Y_{t}= 0.015$m, $Z_{t}= 0.015$m and $\psi_{t} = 6$ pixels. The PID gains used for controlling Y, Z and $\psi$ errors are $k_{py}=68$, $k_{iy}=0.73$, $k_{dy}=4.9$, $k_{pz}=106$, $k_{iz}=0.5$, $k_{dz}=6.1$, $k_{p\psi}=0.8$, $k_{i\psi}=8e^-4$ and $k_{d\psi}=5e-3$.  \par
\begin{figure}[h!]
    \centering
    \includegraphics[width=0.4\textwidth]{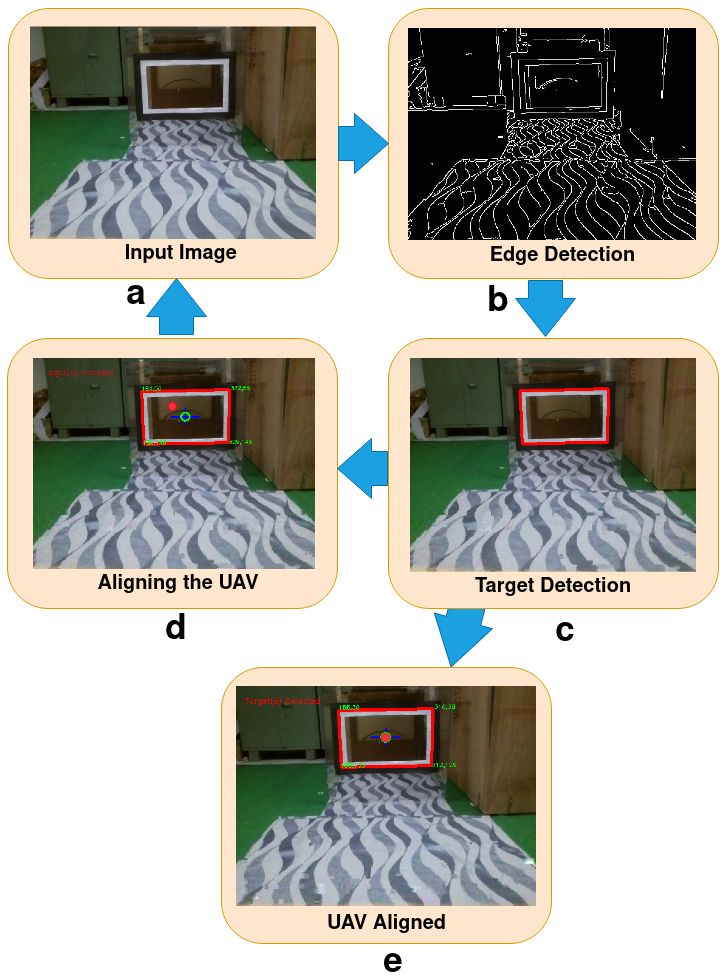}
    \caption{Image processing pipeline for Part I. a) shows the input image from the UAV's camera. b) shows the output of the edge detection algorithm. c) shows the contour of the target window. d) shows the reference point (red dot) and the center of the target window (yellow circle) used for calculating positional error and aligning the UAV. e) shows when the errors are below a threshold and UAV is ready to enter the box.}
    \label{fig:Image flow chart 1}
\end{figure}

The image processing pipeline for Part I is shown in Fig. \ref{fig:Image flow chart 1}. This figure shows how the input image is processed at different stages of the algorithm mentioned in Part I. In this figure, block (a) shows the input image captured of $480$ x $360$ from the UAV's camera. Next, block (b) shows the edge detector output (canny filter). This image is then passed to the border following the algorithm to find the contours. The contour with dimensions resembling the box entrance is selected, as shown in block (c). After calculating the center, corner points, and depth of the target from the selected contour and pinhole camera model, the errors are calculated. In block (d), when the reference point (red dot) does not match the target's center (yellow circle), the UAV is not aligned with the target and needs to adjust its position using the control commands. After minimizing the errors below the thresholds mentioned previously, UAV perfectly aligns with the target, as shown in block (e) of the figure. After this step, the UAV goes inside the box.\par

\begin{figure}[h!]
    \centering
    \includegraphics[width=0.38\textwidth]{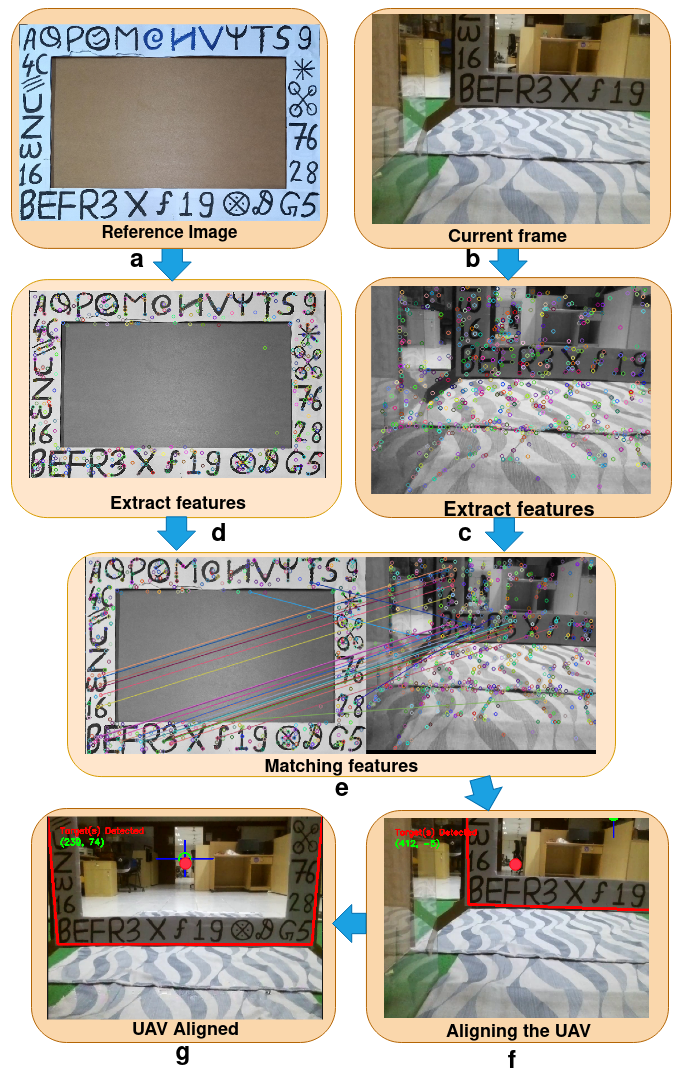}
    \caption{Image processing pipeline for Part II. a) \& b) show the reference image and current image from the camera. c) \& d) shows the features extracted from both images using the SIFT algorithm. e) shows the feature pairs formed by the kNN algorithm. f) shows the partial image of the exit window used for estimating the center point. g) shows the frame in which the UAV is aligned with the target and comes out of the box.}
    \label{fig:Image flow chart 2}
\end{figure}

Figure \ref{fig:Image flow chart 2} shows the image processing pipeline for the algorithm mentioned in Part II. In this figure, block (a) shows the reference image of the exit window of the box, and block (b) shows the current image from the camera. Blocks (c) and (d) show the features extracted from the current and reference images using the Sift algorithm. Features from both images are matched to form the matching pair using the kNN algorithm in block (e). After this, the feature points from the reference image are transformed into the current image using the homography matrix. This transformation gives the approximate estimate of the contour of the exit window from which the center points are estimated. In block f, the UAV is not aligned with the target, so the algorithm adjusts the UAV position by control commands. At last, when the error is below the specified threshold, the UAV is aligned (as shown in block g) and comes out of the box.

\begin{figure}[h!]
    \centering
    \begin{subfigure}{0.35\textwidth}
        \includegraphics[width=\textwidth]{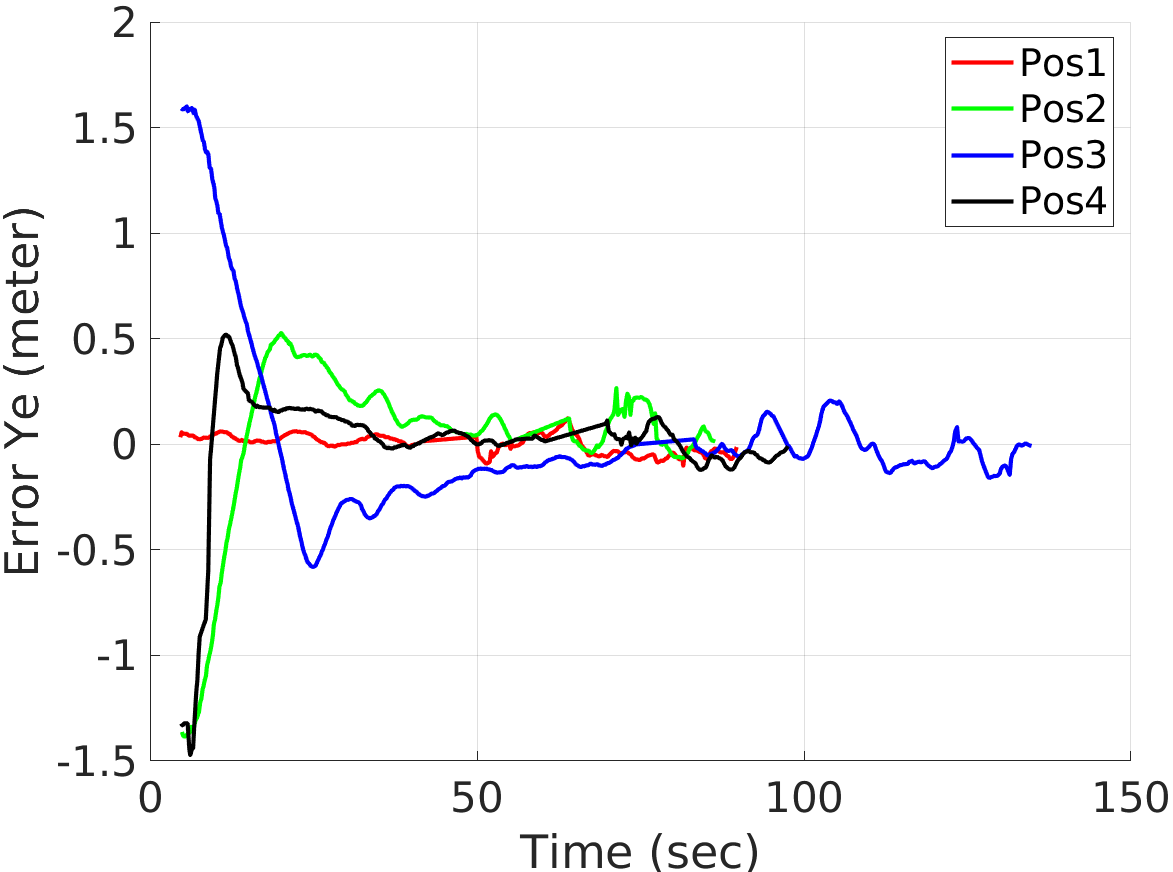}
        \caption{Y-axis error}
        \label{fig:Ye p}
    \end{subfigure}\\
    \begin{subfigure}{0.35\textwidth}
        \includegraphics[width=\textwidth]{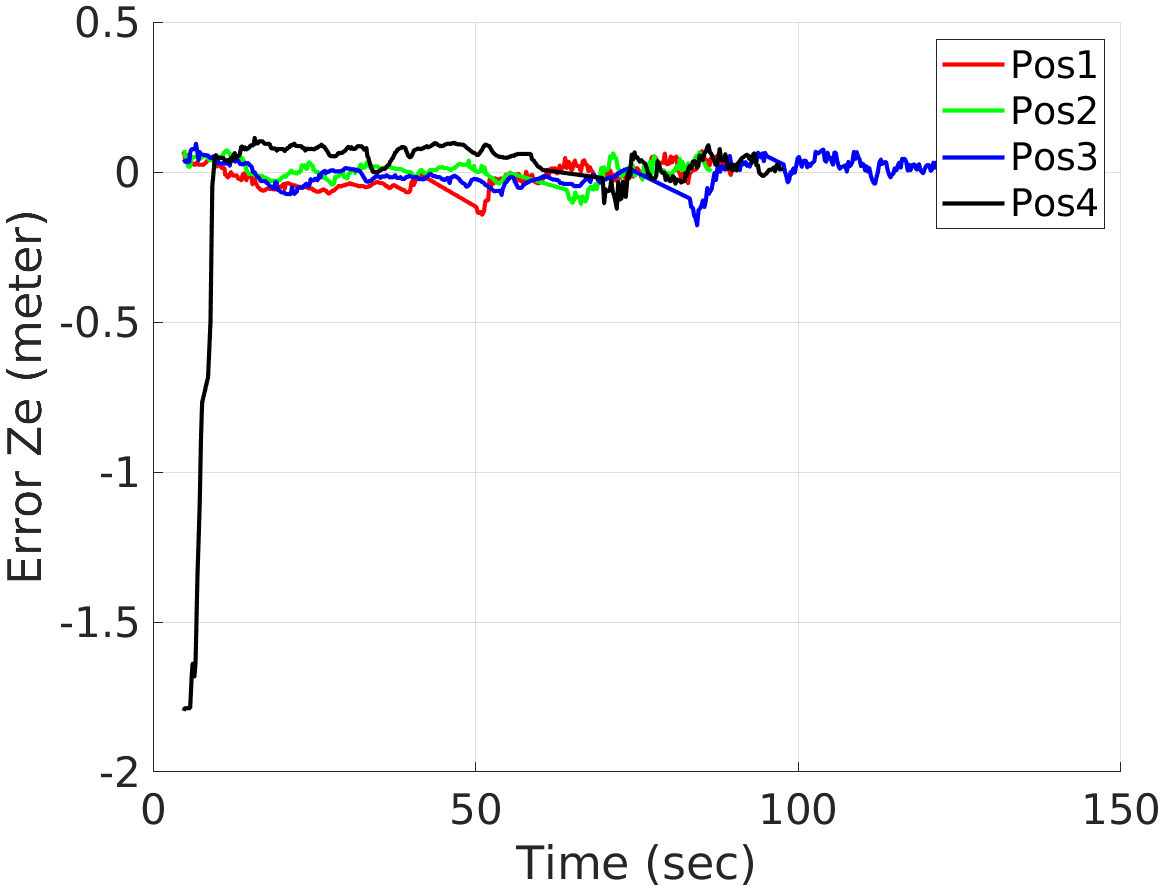}
        \caption{Z-axis error}
        \label{fig:Ze p}
    \end{subfigure}\\
    \begin{subfigure}{0.35\textwidth}
        \includegraphics[width=\textwidth]{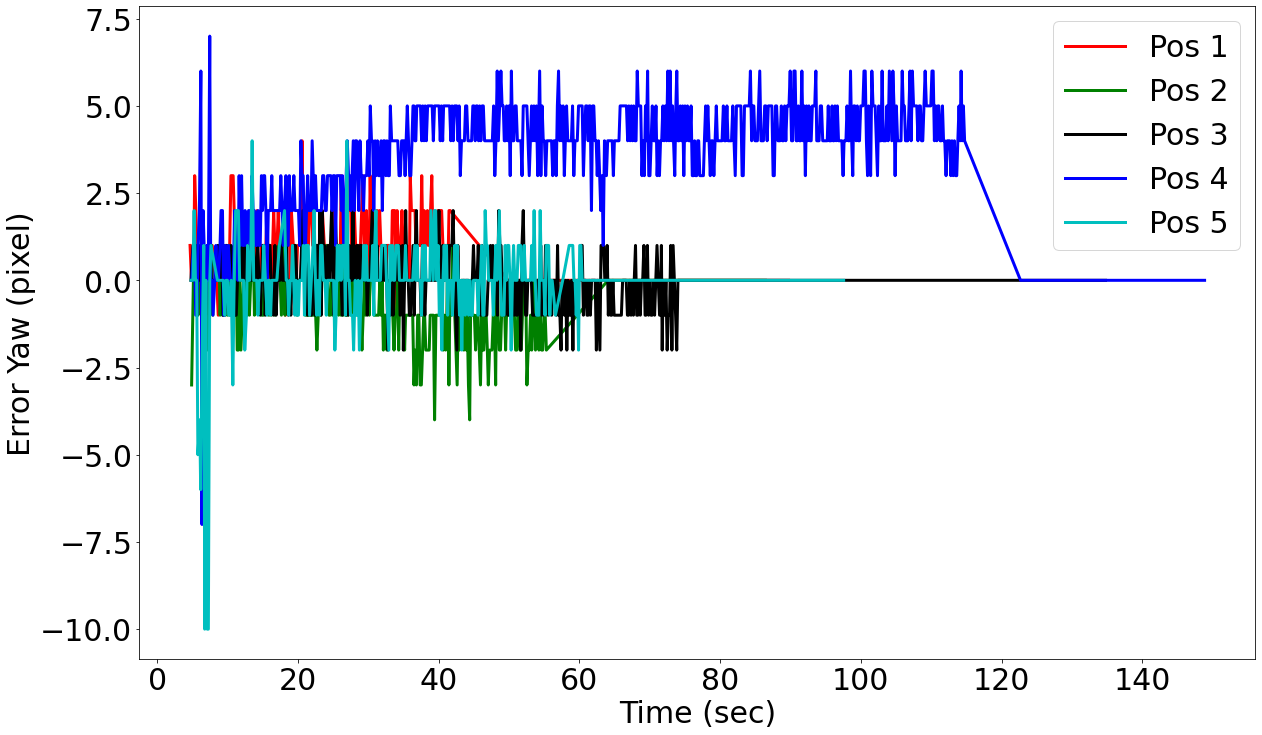}
        \caption{Yaw error}
        \label{fig:Yaw p}
    \end{subfigure}\\
    % \begin{subfigure}{0.4\textwidth}
    %     \includegraphics[width=\textwidth]{Depth Pos.png}
    %     \caption{Depth}
    %     \label{fig:Cal Depth p}
    % \end{subfigure}\\
    \caption{Errors during the flight with different positions and same orientation}
    \label{fig:position flight}
\end{figure}

\begin{figure}[h!]
    \centering
        \includegraphics[width=0.35\textwidth]{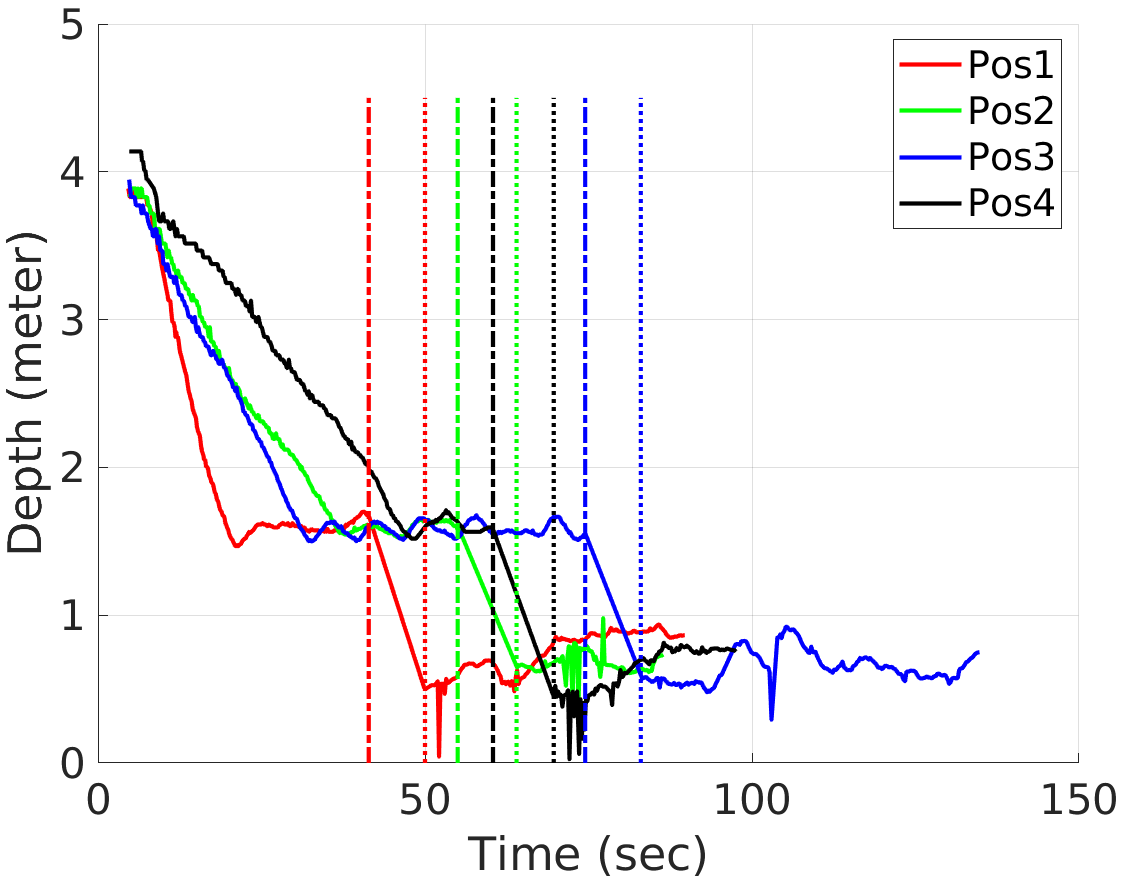}
        \caption{Depth during the flight with different positions and same orientation. The UAV going inside the box is shown by dash-dotted line. The UAV reaching the waypoint inside the box is denoted by dotted line. The UAV flying out of the box is shown by the end of the curve.}
        \label{fig:Cal Depth p}
\end{figure}

\begin{figure}[h!]
    \centering
            \includegraphics[width=0.35\textwidth]{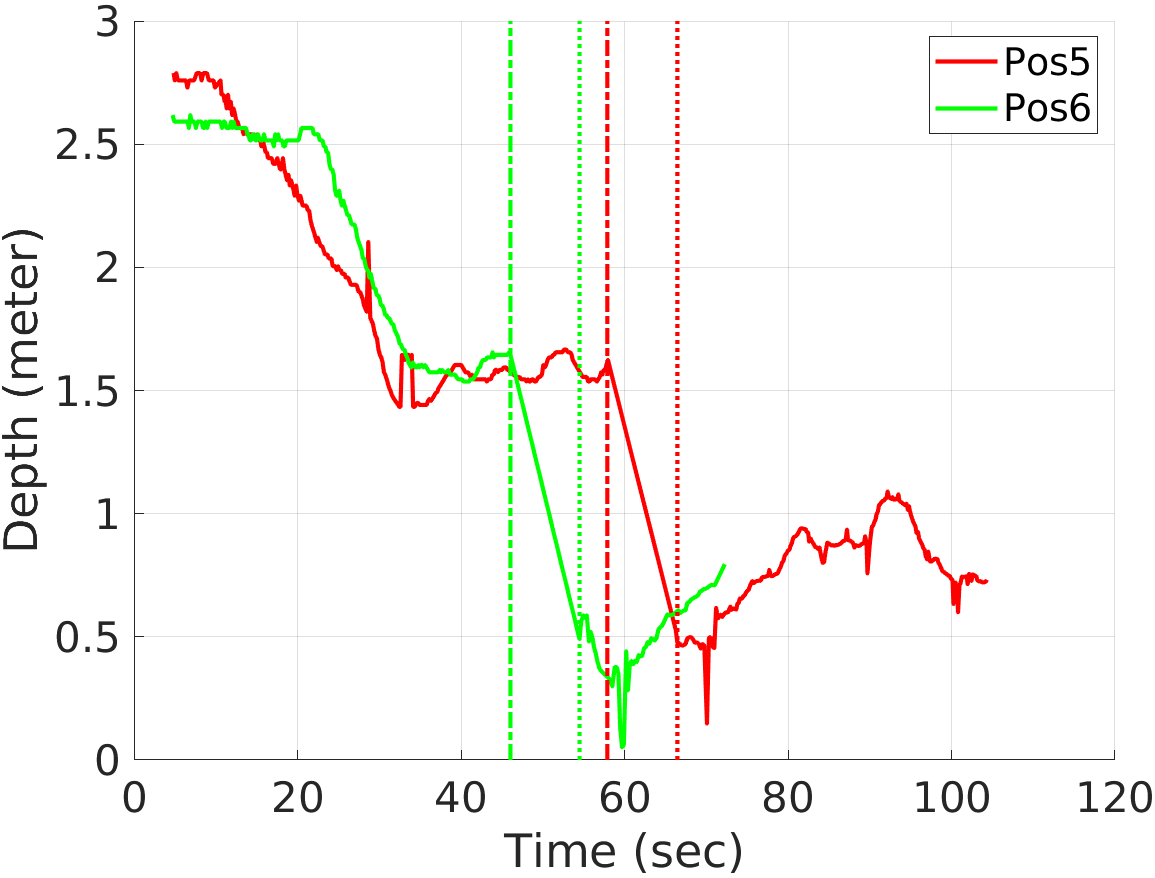}
        \caption{Depth during the flight with same position and different orientation}
        \label{fig:Cal Depth o}
\end{figure}

To test the effectiveness of our algorithms, we used four different initial positions of the UAV. These initial positions are referenced in another world frame shown in Fig \ref{fig:Exp}. These positions are ($0, 0, -0.4$), ($0, 1.5, -0.4$), ($0, -1.5, -0.4)$,  and ($0, 1.5, -2)$. Figure \ref{fig:position flight} shows the positional error (in y and z direction), yaw error, and depth during the flight of the UAV starting from different initial positions. It can be clearly seen that the positional and yaw errors shown in Fig. \ref{fig:Ye p}, \ref{fig:Ye p} \& \ref{fig:Yaw p} are decreasing with time. Depth value shown in Fig. \ref{fig:Cal Depth p} is also decreasing but reaches a local saturation point. This is because when the UAV reaches a value between its threshold, it stops there while continuing to adjust other errors. Once the other errors are below their thresholds, the UAV goes inside the box (depicted by dash-dotted line). After the UAV is inside (shown by dotted line), the value of the depth changes continuously due to aerodynamic disturbances until the UAV flys out of the box (shown by end the curve).    

\begin{figure}[h!]
    \centering
    \begin{subfigure}{0.35\textwidth}
        \includegraphics[width=\textwidth]{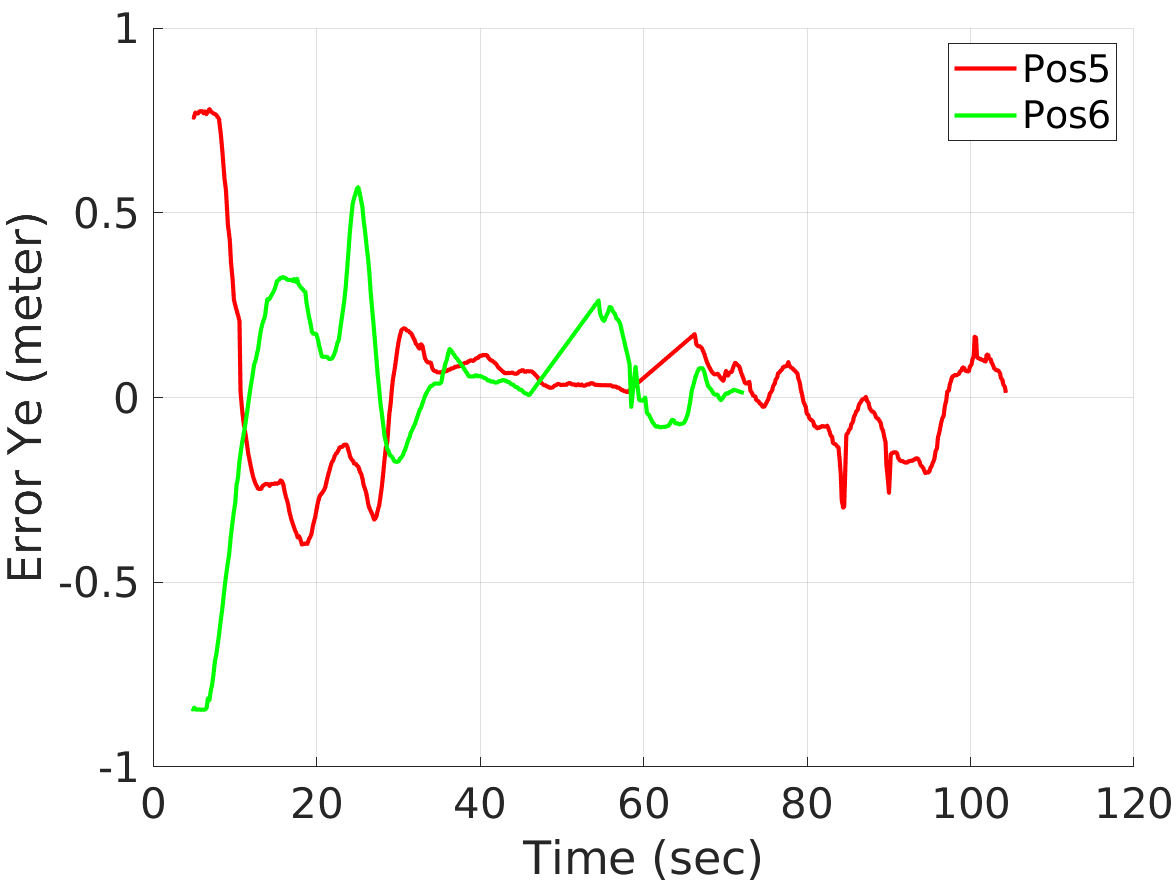}
        \caption{Y-axis error}
        \label{fig:Ye o}
    \end{subfigure}\\
    \begin{subfigure}{0.35\textwidth}
        \includegraphics[width=\textwidth]{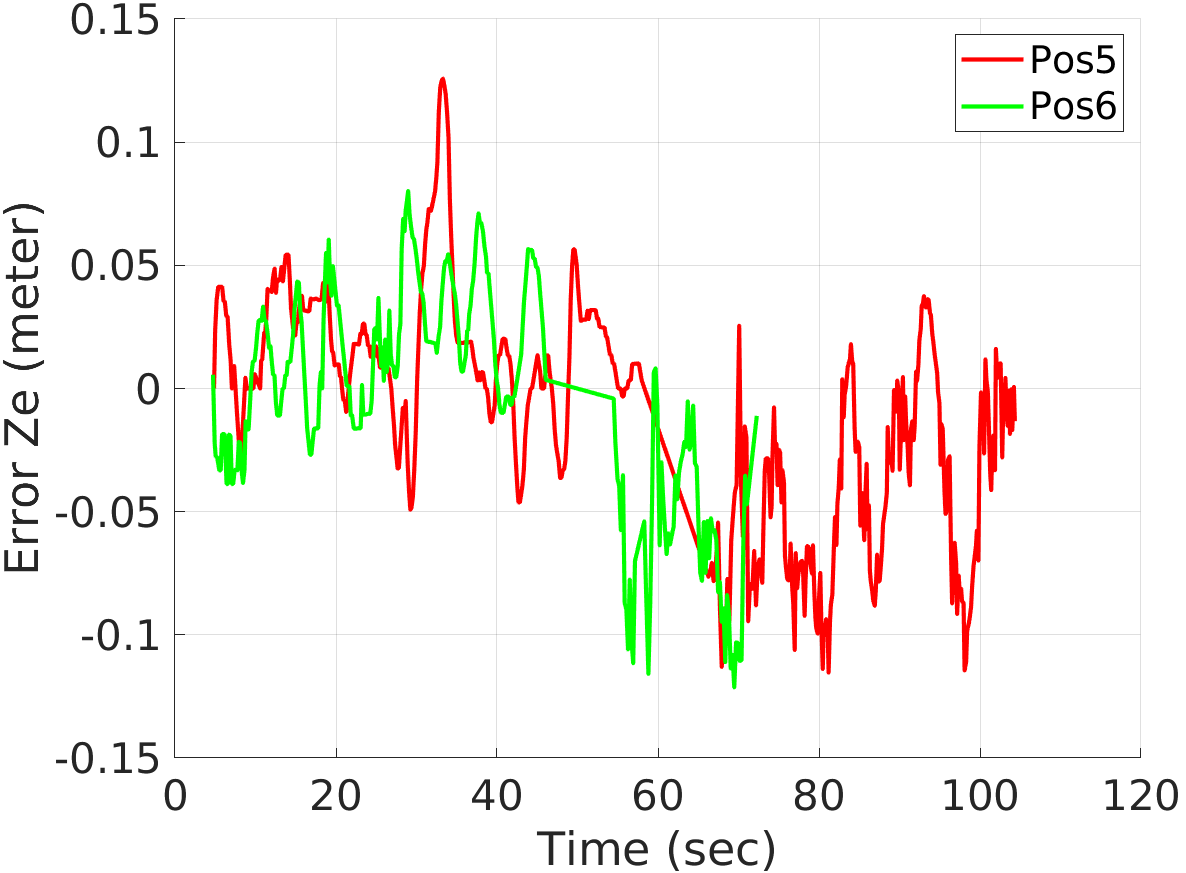}
        \caption{Z-axis error}
        \label{fig:Ze o}
    \end{subfigure}\\
    \begin{subfigure}{0.35\textwidth}
        \includegraphics[width=\textwidth]{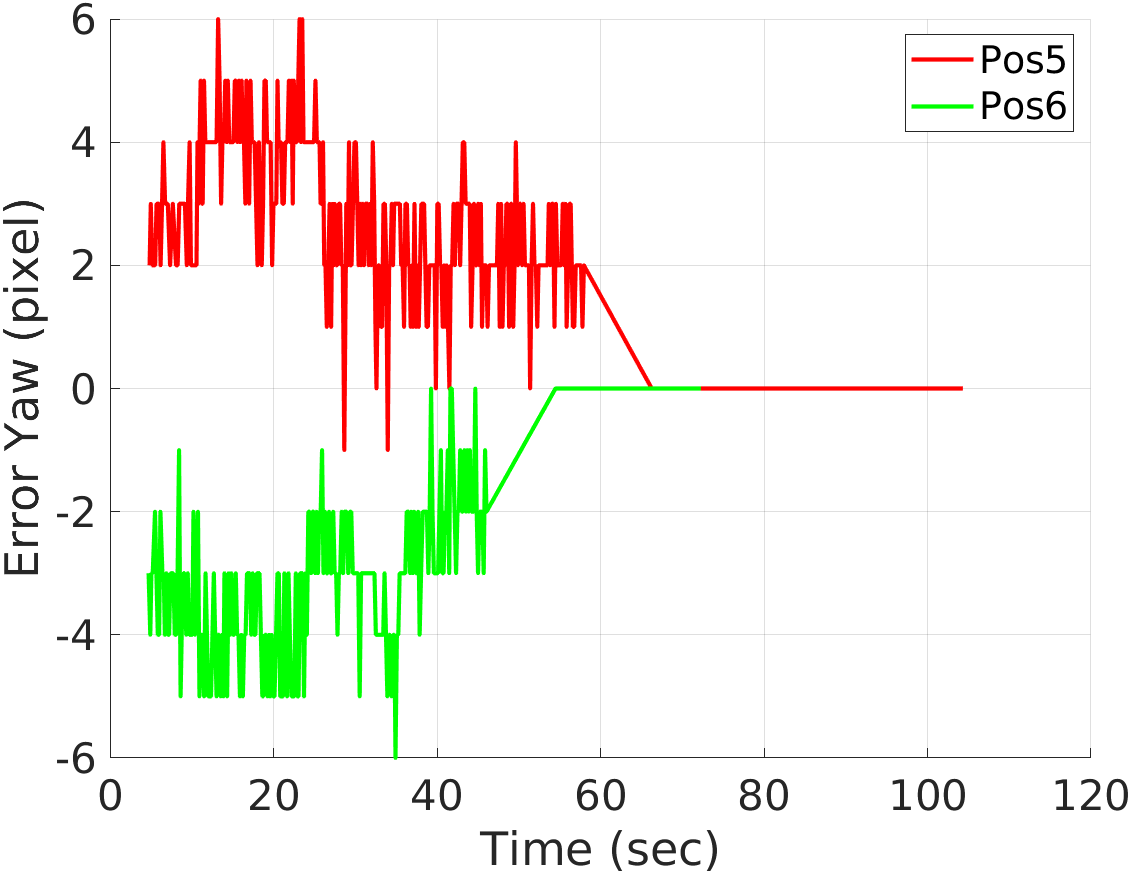}
        \caption{Yaw error}
        \label{fig:Yaw o}
    \end{subfigure}\\
    \caption{Errors during the flight with same position and  different orientation}
    \label{fig:orientation flight}
\end{figure}

Additionally, we used two different initial orientations with the same position to test our algorithm further. The initial position used for this case is $(1.25, 0, -0.4)$ and the initial orientations used are $(0, 0, -20^{\circ})$ and $(0, 0, 20^{\circ})$. Figure \ref{fig:Ye o}, \ref{fig:Ze o} \& \ref{fig:Yaw o} shows the positional and yaw error. It is evident from the plots that the error values are decreasing with time for this case also. Moreover, the Depth shown in Fig. \ref{fig:Cal Depth o} also shows the same behavior but instead of having two saturation points, it shows two troughs. This could be because this case takes less time to stabilize its position errors due to its initial position.

We do not control the yaw once the UAV is inside the box because the homography technique can sometimes produce erroneous position estimates. As a result, we do not rely on homography for yaw adjustment. Consequently, the yaw error for both cases (depicted in Fig. \ref{fig:Yaw p} and \ref{fig:Yaw o}) becomes zero once the UAV enters the box. Also, the yaw error is in pixels which are integer values, this is the reason the yaw error graphs for both the cases are not smooth like the others.

\section{CONCLUSION}
In this study, we developed a novel vision-based control algorithm for UAV navigation through narrow passages by combining segmentation, homography, and PID controller. The effectiveness of this algorithm is demonstrated with several collision-free flights through a small rectangular box under varying initial conditions. Notably, our algorithm navigated the UAV in and out of the box in a single continuous flight without colliding with the walls, even when visual feedback was partially occluded. Future research will focus on using machine learning to increase the robustness of our algorithm with different lighting conditions.

% \addtolength{\textheight}{-12cm} 
\bibliographystyle{IEEEtran}
\bibliography{IEEEabrv, references}

\end{document}